\documentclass[11pt,a4paper]{article}
\usepackage[hyperref]{acl2021}
\usepackage{times}
\usepackage{latexsym}

\usepackage{CJK}
\usepackage{enumitem}
\usepackage{float}
\usepackage{lineno,hyperref}
\usepackage{verbatim}
\usepackage{booktabs}
\usepackage{multirow}
\usepackage{makecell}
\usepackage{algorithm}
\usepackage{algorithmic}
\usepackage{graphicx}
\usepackage{microtype}
\usepackage{amsmath}%数学符号  
\usepackage{bm}%专门处理数学粗体的bm宏包
\usepackage{enumerate}
\usepackage{multirow}
\usepackage{amssymb}
\usepackage{setspace}
\usepackage{array}
\usepackage{tabu}
\usepackage{adjustbox}
\usepackage{tablefootnote}
\usepackage[normalem]{ulem}
\usepackage{caption}
\usepackage{afterpage}
\usepackage{color}

\usepackage{microtype}
\aclfinalcopy

\title{Exploration and Exploitation: Two Ways to Improve\\Chinese Spelling Correction Models}

\author{Chong Li\footnotemark[2] , Cenyuan Zhang\footnotemark[2] , Xiaoqing Zheng\footnotemark[1] ,  Xuanjing Huang\\
  School of Computer Science, Fudan University, Shanghai, China\\
  Shanghai Key Laboratory of Intelligent Information Processing \\
  \texttt{\{chongli17, cenyuanzhang17, zhengxq, xjhuang\}@fudan.edu.cn} \\
}
\date{}

\begin{document}
\begin{CJK}{UTF8}{gbsn}

\maketitle

\renewcommand{\thefootnote}{\fnsymbol{footnote}} %将脚注符号设置为fnsymbol类型，即特殊符号表示
\footnotetext[2]{These authors contributed equally to this work.} %对应脚注[1]
\footnotetext[1]{Corresponding Author}

\renewcommand{\thefootnote}{\arabic{footnote}}

\begin{abstract}
A sequence-to-sequence learning with neural networks has empirically proven to be an effective framework for Chinese Spelling Correction (CSC), which takes a sentence with some spelling errors as input and outputs the corrected one. 
% However, the performance of CSC models with this framework heavily relies on the size and quality of corpus.
However, CSC models may fail to correct spelling errors covered by the confusion sets, and also will encounter unseen ones. 
We propose a method, which continually identifies the weak spots of a model to generate more valuable training instances, and apply a task-specific pre-training strategy to enhance the model. The generated adversarial examples are gradually added to the training set.
% We propose a method inspired by adversarial training to generate more valuable training instances by continually identifying the weak spots of a model, and to enhance the model by gradually adding the generated adversarial examples to the training set.
Experimental results show that such an adversarial training method combined with the pre-training strategy can improve both the generalization and robustness of multiple CSC models across three different datasets, achieving state-of-the-art performance for CSC task.\footnote{The source codes are available at \href{https://github.com/FDChongli/TwoWaysToImproveCSC}{https://github.com/\\FDChongli/TwoWaysToImproveCSC}.}
\end{abstract}

\section{Introduction}
Chinese Spelling Correction (CSC) aims to detect and correct spelling mistakes in Chinese texts. Many Chinese characters are visually or phonologically similar, while their semantic meaning may differ greatly. 
Spelling errors are usually caused by careless writing, automatic speech recognition, and optical character recognition systems.
The CSC task has received steady attention over the past two decades \citep{chang1995new,xin-etal-2014-improved,AutoGen-2018-EMNLP,hong-etal-2019-faspell}.
Unlike English, Chinese texts are written without using whitespace to delimit words, and it is hard to identify whether and which characters are misspelled without the information of word boundaries.
The context information should be taken into account to reconstruct the word boundaries when correcting spelling mistakes, which makes CSC a long-standing challenge for Chinese NLP community. 
% As an important preprocessing step for many downstream tasks such as word segmentation, named entity recognition, and essay scoring, CSC has been 

Many early CSC systems follow the same recipe with minor variations, adopting a three-step strategy: detect the positions of spelling errors; generate candidate characters for these positions; and select a most appropriate one from the candidates % for each position 
to replace the misspelling \citep{yeh-etal-2013-chinese,yu-li-2014-chinese,zhang-etal-2015-hanspeller,wang-etal-2019-confusionset}.
Recently, a sequence-to-sequence (seq2seq) learning framework with neural networks has empirically proven to be effective for CSC, which transforms a sentence with errors to the corrected one \cite{zhang-etal-2020-spelling,cheng-etal-2020-spellgcn}.
However, even if training a CSC model with the seq2seq framework normally requires a huge amount of high-quality training data, it is still unreasonable to assume that all possible spelling errors have been covered by the \emph{confusion} sets (i.e. a set of characters and their visually or phonologically similar characters which can be potentially confused) extracted from the training samples.
New spelling errors occur everyday.
A good CSC model should be able to exploit what it has already seen in the training instances in order to achieve reasonable performance on easy spelling mistakes, but it can also explore in order to generalize well to possible unseen misspellings.  

In this study, we would like to pursue both the \emph{exploration} (unknown misspellings) and \emph{exploitation} (the spelling errors covered by the confusion sets) when training the CSC models.
To encourage a model to explore unknown cases, we propose a character substitution-based method to pre-train the model.
The training data generator chooses about $25\%$ of the character positions at random for prediction. If a character is chosen, we replace it with the character randomly selected from its confusion set ($90\%$ of the time) or a random character ($10\%$ of the time). Then, the model is asked to predict the original character. 

% Even though the confusion sets are given and fixed, it is still infeasible to explore all possible
% combinations in which each character in a sentence can be replaced with any one from its confusion set.
Because of the combination of spelling errors and various contexts in which they occur, even though the confusion sets are given and fixed, models may still fail to correct characters that are replaced by any character from its confusion set.
To better exploit what the models has experienced during the training phase, we generate more valuable training data via adversarial attack (i.e. tricking models to make false prediction by adding imperceptible perturbation to the input  \citep{szegedy2014intriguing}), targeting the weak spots of the models, which can improve both the quality of training data for fine-tuning the CSC models and their robustness against adversarial attacks.
Inspired by adversarial attack and defense in NLP \citep{jia-liang-2017-adversarial,zhao2018generating,Cheng_Yi_Chen_Zhang_Hsieh_2020,wang-zheng-2020-improving}, we propose a simple but efficient method for adversarial example generation: we first identify the most vulnerable characters with the lowest generation probabilities estimated by a pre-trained model, and replace them with characters from their confusion sets to create the adversarial examples.

% An adversarial example generation algorithm is designed, which identifies vulnerable positions in a sentence based on output logit and exploits knowledge of the confusion set by substituting characters with candidates from it. Adversarial examples generated by this algorithm were used in adversarial training. To help the model to explore, we pre-train these models with a large corpus generated by substituting randomly chosen characters in clean sentences. Clean sentences were collected from Wikipedia and Weibo corpora and filtered by a base CSC model.
Once the adversarial examples are obtained, they can be merged with the original clean data to train the CSC models.
The examples generated by our method are more valuable than those already existed in the training set because they are generated towards to the weak spots of the current models.
Through extensive experimentation, we show that such adversarial examples can improve both generalization and robustness of CSC models.
If a model pre-trained with our proposed character substitution-based method is further fine-tuned by adversarial training, its robustness can be improved about $3.9\%$ while without suffering too much loss (less than $1.1\%$) on the clean data.

% In conclusion, our solution consists of two aspects, exploration and exploitation, to improve both the generalization and robustness of CSC models. Pre-training is used to cover unknown errors, while adversarial training is used to identify the weak spots of a current model and gradually make up for the weak spots founded. Through pre-training and adversarial training, models showed better performance and robustness on benchmark datasets.

\section{Method}

\subsection{Problem Definition}
Chinese Spelling Correction aims to identify  incorrectly used characters in Chinese texts and giving its correct version.  
Given an input Chinese sentence $X=\{x_1,...,x_n\}$ consisting of $n$ characters, which may contain some spelling errors, the model takes $X$ as input and outputs an output sentence $Y=\{y_1,...,y_n\}$, where all the incorrect characters are expected to be corrected. 
This task can be formulated as a conditional generation problem by modeling and maximizing the conditional probability of $P(Y|X)$.

\subsection{Base Models}
We use vanilla BERT \citep{devlin-etal-2019-bert} and two recently proposed BERT-based models  \citep{cheng-etal-2020-spellgcn,zhang-etal-2020-spelling} as our base models. 
When applying BERT to the CSC task, the input is a sentence with spelling errors, and the output representations are fed into an output layer to predict target tokens. We tie the input and output embedding layer, and all the parameters are fine-tuned using task-specific corpora.
Soft-Masked BERT \cite{zhang-etal-2020-spelling} uses a Bi-GRU network to detect errors, and applies a BERT-based network to correct errors. SpellGCN \cite{cheng-etal-2020-spellgcn} utilizes visual and phonological similarity knowledge through a  specialized graph convolutional network and substitutes parameters of the output layer of BERT with the final output of it.
% Input a Chinese sentence, which may contain spelling errors, to BERT. For each character, probability to predict a target character is computed using the hidden state of the last layer of BERT. Soft-Masked BERT \cite{zhang-etal-2020-spelling} consists of a Bi-GRU network and a network based on BERT to detect and correct errors, respectively. SpellGCN \cite{cheng-etal-2020-spellgcn} makes use of character embeddings of BERT and gains representations containing visual and phonological similarity knowledge to predict target characters. 
% Input a Chinese sentence, which may contain spelling errors, to BERT and gain hidden states of its last layer, denoted as $H=\{h_1,...,h_n\}$. For each position $i$, probability used to predict a target character is computed by
% \begin{equation}
%     P_i=softmax(Wh_i+b)
% \end{equation}
% where $W\in \mathbb{R}^{V \times E}$ and $b\in \mathbb{R}^{V}$ are trainable parameters of the model. $E$ is the embedding size and $V$ is the size of vocabulary. The character with the largest probability in the vocabulary will be output

These models achieved state-of-the-art or close to state-of-the-art performance on the CSC task.  
However, we found that their performance and robustness could be further improved through pre-training and adversarial training, which help models explore unseen spelling errors and exploit weak points of themselves.
% But through exploration and exploitation, namely pre-training using generated large corpus and adversarial training using examples generated by adversarial algorithm, we can further improve their performance and robustness.  

\subsection{Pre-training Method}
We collected unlabeled sentences from Wikipedia and Weibo corpora \cite{shang-etal-2015-neural}, covering both formal and informal Chinese texts. 
Training example pairs are generated by substituting characters in clean sentences, and models are trained to predict the original character. 
According to \citet{chen2011improve}, a sentence contains no more than two spelling errors on average, so we select and replace $25\%$ characters in a sentence. 
The chosen Chinese character will be substituted by a character randomly selected from its confusion set ($90\%$ of the time) or a random Chinese character ($10\%$ of the time). 
The latter helps models to explore unknown misspellings not covered by the confusion sets. 
% About nine million examples are generated through this approach for pre-training.

\subsection{Adversarial Example Generation and Adversarial Training}

% \subsubsection{Adversarial Attack Algorithm}

% We proposed an adversarial attack algorithm to identify the weak spots of trained CSC models by replacing the tokens in a sentence with the spelling mistakes people may make.

To efficiently identify and alleviate the weak spots of trained CSC models, we designed an adversarial attack algorithm for CSC tasks,  which replaces the tokens in a sentence with spelling mistakes.

The adversarial examples generation algorithm in this paper can be divided into two main steps: (1) determine the vulnerable tokens to change (2) replace them with the spelling mistakes that most likely to occur in the contexts (Algorithm \ref{alg:generate}).

For the i-th position of input sentence $X$, the positional score $s_i$ can be obtained by the logit output $o_i$ as follows:
\begin{equation}
    s_i=o_i^{y_i}-o_i^{m_i} (o_i^{m_i}=max\{o_i^{r},r \neq y_i\})
\end{equation}

where $o_i^r$ denotes the logit output of character $r$ in the i-th position, and $y_i$ denotes the i-th character of ground truth sentence $Y$. The lower the positional score, the less confident the model is in predicting the position. Attacking this position makes the model output more likely to change. Once the positional score of each character in the input sentence is calculated, we sort these positions in ascending order according to the positional scores. This process can reduce the substitutions and maintain the original semantics as much as possible.

Once a vulnerable position is determined, the token at that position is replaced with one of its phonologically or visually similar characters. Confusion set $D$ contains a set of visually or phonologically similar characters. In order to fool the target CSC model while maintaining the context, the character with the highest logit output in the confusion set is used as a replacement.

Given a sentence in training sets, its adversarial examples are generated by substituting a few characters based on the algorithm mentioned above. 
 Adversarial training was conducted with these examples, improving the robustness of CSC models by alleviating their weak spots, and exploiting knowledge about easy spelling mistakes from confusion sets to help models generalize better.

\begin{algorithm}[htb] 
\small

\caption{Adversarial Attack Algorithm} 
\label{alg:generate}
\begin{algorithmic}[1] %这个1 表示每一行都显示数字
% \REQUIRE ~~\\ %算法的输入参数：Input
% 	$X=\{x_1,x_2,...,x_n\}$, input Chinese sentence; \\
% 	$Y=\{y_1,y_2,...,y_n\}$, the corresponding ground truth;\\
% 	$\lambda$, proportion of characters can be changed;\\
% 	$f$, a target CSC model; \\
% 	$D$, a confusion set created based on visually or phonologically similar characters;
% \ENSURE ~~\\ %算法的输出：Output
% $\widehat{X}=\{\hat{x}_1,\hat{x}_2,...,\hat{x}_n\}$, adversarial example;
% \STATE $\widehat{X} \gets X$ \
% \IF{$f(X)\neq Y$}
% 	\STATE return $\widehat{X}$
% \ELSE
% 	\STATE $O=\{o_1,o_2,...,o_n\} \gets $ Logit output of $f(X)$ \
% 	\STATE $P=\{p_1,p_2,...,p_k\} \gets$ Sort the position $p_i$ in ascending order based on $s_{p_i}\ (\ 1\le p_i\le n,\ y_{p_i} $ is a Chinese character and  $o_{p_i}^{m_{p_i}}=max\{o_{p_i}^{r},r \neq y_{p_i}\}\ )$\
% % 	$s_{p_i}=(o_{p_i}^{y_{p_i}}-o_{p_i}^{m_{p_i}})$ \
% % 	\STATE $(1\le p_i\le n,y_{p_i} $ is a Chinese character and  $o_{p_i}^{m_{p_i}}=max\{o_{p_i}^{r},r \neq y_{p_i}\})$\
% 	\STATE $num\gets 0$\
% 	\FOR{each $i \in [1,k]$}
% 		\IF{$x_{p_i}\neq y_{p_i}$}
% 			\STATE {\bf continue} \
% 		\ENDIF
% 		\STATE $\hat{x}_{p_i}\gets m_{p_i}$, where $m_{p_i} \in D(x_{p_i})\ $and$\ p^{m_{p_i}}_i=max\{p^{r}_i,r\in D(x_{p_i})\}$ \
% 		\STATE $num\gets num+1$ \
% 		\IF {$f(\widehat{X})\neq Y\  \| \ num> \lambda \cdot n$}
% 			\STATE return $\widehat{X}$\
% 		\ENDIF
% 	\ENDFOR
% \ENDIF
% \RETURN $\widehat{X}$; %算法的返回值
\REQUIRE ~~\\
$X = \left\{ x_{1},x_{2},\ldots,x_{n} \right\}$, input Chinese sentence;\\
$Y = \left\{ y_{1},y_{2},\ldots,y_{n} \right\}$, the corresponding ground truth;\\
$\lambda$, proportion of characters can be changed;\\
$f$, a target CSC model;\\
$D$, a confusion set created based on visually or phonologically similar characters;
\ENSURE ~~\\
$\widehat{X} = \left\{ \hat{x}_{1},\hat{x}_{2},\ldots,\hat{x}_{n} \right\}$, adversarial example;
\STATE $\left. \widehat{X}\gets X \right.$
\IF{$f\left( X \right) \neq Y$}
  \RETURN $\widehat{X}$
\ELSE
  \STATE $\left. num\leftarrow 0 \right.$
  \WHILE{$f(\widehat{X}) = Y\ \&\&\ num \leq \lambda \cdot n$ }
    \STATE $O=\{o_1,o_2,...,o_n\} \gets $ Logit output of $f(\widehat{X})$ \
    \STATE $P=\{p_1,p_2,...,p_k\} \gets$ Sort the position $p_i$ in ascending order based on $s_{p_i}\ (\ 1\le p_i\le n$ and $y_{p_i} $ is a Chinese character\ )\
    \FOR{each $i \in [1,k]$}
      \IF{$\hat{x}_{p_{i}} \neq y_{p_{i}}$}
        \STATE {\bf continue}
      \ENDIF
    \STATE $\left. \hat{x}_{p_{i}}\gets m_{p_{i}} \right.$, where $m_{p_{i}} \in D\left( x_{p_{i}} \right)$ and $p_{i}^{m_{p_{i}}} = max\left\{ p_{i}^{r},p_{i}^{r} \in D\left( x_{p_{i}} \right) \right\}$
    \STATE {\bf break}
    \ENDFOR
  \STATE $\left. num\gets num + 1 \right.$
  \ENDWHILE
\ENDIF
\RETURN $\widehat{X}$
\end{algorithmic}

\end{algorithm}

\section{Experiments}
\subsection{Datasets}
Statistics of the datasets used are shown in Table \ref{tab:data}.
\paragraph{Pretraining data}  We generated a large corpus by a character substitution-based method. Models were first pre-trained on these nine million sentence pairs, and then fine-tuned using the training data mentioned below.
\paragraph{Training data}  The training data contained three human-annotated training datasets, SIGHAN 2013 \citep{wu-etal-2013-Bake-off}, SIGHAN 2014 \citep{yu-etal-2014-Bake-off}, and SIGHAN 2015  \citep{tseng-etal-2015-Bake-off}. We also utilized an automatically generated dataset \cite{AutoGen-2018-EMNLP}. 
\begin{table}[H]
\small
\caption{Statistics information on the used data resources. A subset of the Wikipedia corpus and Weibo corpus, denoted by Wikipedia$^*$ and Weibo$^*$ respectively, was sampled from the entire corpus.}
\vspace{-0.2cm}
\centering

\begin{tabular}{llrr}
\toprule
\multicolumn{2}{l}{Pre-Training Data\ \ \ \ \ \ \ \ \ \ \ \ \ \ \ \ \ \ } & \#Line & Avg. Length \\
\midrule
\multicolumn{2}{l}{Wikipedia$^*$} & 4,531,007 & 40.2\\
\multicolumn{2}{l}{Weibo$^*$} & 4,770,015 & 16.3\\
\bottomrule
\end{tabular}
\begin{tabular}{lrrr}
\toprule
Training Data & \#Line & Avg. Length & \#Errors\\
\midrule
\citep{AutoGen-2018-EMNLP} & 271,329 & 42.6 & 381,962\\
SIGHAN 2013 & 350 & 49.3 & 339 \\
SIGHAN 2014 & 3,437 & 49.6 & 5,136 \\
SIGHAN 2015 & 2,339 & 31.3 & 3,048 \\
\bottomrule
\toprule
Test Data & \#Line & Avg. Length & \#Errors\\
\midrule
SIGHAN 2013 & 1,000 & 74.3 & 1,221 \\
SIGHAN 2014 & 1,062 & 50.0 & 771 \\
SIGHAN 2015 & 1,100 & 30.7 & 705 \\
\bottomrule
\end{tabular}
\label{tab:data}
\end{table}
\vspace{-0.3cm}
\begin{table*}[htp]

\caption{\label{tab:performance} Performance of three models trained with the proposed pretraining strategy and adversarial training method. ``\textbf{CLEAN}'' stands for the testing results on the clean data, and ``\textbf{ATTACK}'' denotes the F1 scores under test-time attacks.
``\textbf{DET}'' and ``\textbf{COR}'' denote the F1 scores of detection and correction. The F1 scores were increased $4.1\%$ on average by our pre-training method across the various models on the different datasets. Models' robustness was also improved about $3.9\%$ while without suffering too much loss (less than $1.1\%$) on the clean data.
}
\centering
\small

\setlength{\tabcolsep}{2mm}
 \begin{tabu}{lcccccccccccc}
 
 \toprule[1.2pt]
  & \multicolumn{4}{c}{\textbf{ SIGHAN-2013}} & \multicolumn{4}{c}{\textbf{SIGHAN-2014}} &    \multicolumn{4}{c}{\textbf{SIGHAN-2015}} \\
  \cmidrule(r){2-5}  \cmidrule(r){6-9} \cmidrule(r){10-13} \noalign{\smallskip}
  & \multicolumn{2}{c}{\textbf{CLEAN}} & \multicolumn{2}{c}{\textbf{ATTACK}} & \multicolumn{2}{c}{\textbf{CLEAN}} & \multicolumn{2}{c}{\textbf{ATTACK}}  & \multicolumn{2}{c}{\textbf{CLEAN}} & \multicolumn{2}{c}{\textbf{ATTACK}} \\ 
\cmidrule(r){2-3}  \cmidrule(r){4-5} \cmidrule(r){6-7} \cmidrule(r){8-9} \cmidrule(r){10-11} \cmidrule(r){12-13} \noalign{\smallskip}
\multicolumn{1}{c}{\textbf{Model}}& \textbf{DET}&\textbf{COR}& \textbf{DET}&\textbf{COR}& \textbf{DET}&\textbf{COR} & \textbf{DET}&\textbf{COR}& \textbf{DET}&\textbf{COR}& \textbf{DET}&\textbf{COR}\\
   \midrule[0.8pt]
 BERT    & $82.9$ & $82.1$ & $33.6$  &  $15.8$ & $66.8$ &  $65.0$ & $41.7$ & $19.0$ & $76.3$ &  $74.4$ & $25.1$ &  $13.7$ \\ 
 \specialrule{0em}{0pt}{0pt}
 
 \scriptsize \ \ \ +\ Pre-trained for CSC & $\bf 84.9$ & $\bf 84.4$ & $48.5$ &  $29.6$ & $\bf 70.4$ &  $\bf 68.6$ & $51.4$ & $32.4$ & $ 79.8$ &  $ 78.0$ & $39.0$ &  $26.9$ \\ 
  \scriptsize \ \ \ +\ Adversarial training & $84.0$ &  $83.5$ & $\bf 50.8$ &  $\bf 31.3$ & $ 68.4$ & $ 66.8$ & $\bf 54.9$ & $\bf 38.0$ & $\bf 80.0$ &  $\bf 78.2$ & $\bf 45.9$ &  $\bf36.0$ \\ 
  \midrule[0.8pt]

 SpellGCN    & $80.8$ & $80.0$ & $25.6$ &  $22.6$ & $64.8$ &  $63.6$ & $29.0$ & $24.3$ & $73.6$ &  $71.5$ & $18.8$ &  $17.4$ \\  
 \specialrule{0em}{0pt}{0pt}
 \scriptsize \ \ \ +\ Pre-trained for CSC & $\bf 84.6$ & $\bf 84.0$ & $28.8$ &  $25.8$ & $\bf 67.3$ &  $\bf 66.4$ & $35.4$ & $27.1$ & $79.6$ &  $77.7$ & $26.2$ &  $25.2$ \\ 
  \scriptsize \ \ \ +\ Adversarial training & $83.4$ & $82.6$ & $\bf 30.2$ &  $\bf 26.0$ & $66.4$ &  $65.4$ & $\bf 35.9$ & $\bf 29.5$ & $\bf 79.6$ &  $\bf 77.8$ & $\bf 28.2$ &  $\bf 25.2$ \\ 
  \midrule[0.8pt]

 Soft-masked BERT    & $80.6$ & $79.1$ & $27.7$ &  $4.0$ & $62.2$ &  $59.6$ & $29.8$ & $7.1$ & $72.4$ &  $69.6$ & $15.5$ &  $5.3$ \\  
 \specialrule{0em}{0pt}{0pt}
 \scriptsize \ \ \ +\ Pre-trained for CSC &  $\bf 84.9$ & $\bf 84.2$ &  $27.3$ &  $6.0$ & $\bf 67.2$ &  $\bf 65.6$ & $30.7$ & $8.6$ & $\bf 77.2$ &  $\bf 74.5$ & $22.2$ &  $6.5$\\ 
  \scriptsize \ \ \ +\ Adversarial training & $84.1$ &  $83.3$ & $\bf 32.5$ &   $\bf 8.1$ & $65.0$ & $62.7$ & $\bf 40.5$ & $\bf 13.4$ & $ 76.2$ &  $ 73.8$ & $\bf 30.3$ &  $\bf 11.4$\\ 

\bottomrule[1.2pt]
\end{tabu}
\end{table*}
\paragraph{Test data} Models' performance in detection and correction stage was evaluated in sentence level on three benchmark datasets, in the metrics of F1 scores (detection and correction). Characters in these datasets were transferred into simplified Chinese characters using OpenCC\footnote{https://github.com/BYVoid/OpenCC}. We revised the processed datasets for one simplified Chinese character may correspond to multiple traditional Chinese characters.

\subsection{Models and Hyper-parameter Settings}

% \input{attackPerformance}
% Our training method was applied on BERT and two state-of-the-art models on the CSC task. Soft-Masked BERT\cite{zhang-etal-2020-spelling} consists of a Bi-GRU network and a network based on BERT to detect and correct errors, respectively. SpellGCN\cite{cheng-etal-2020-spellgcn} makes use of character embeddings of BERT and gains representations containing visual and phonological similarity knowledge. 

For BERT and Soft-Masked BERT, we used the BERT model pre-trained on Chinese text provided by transformers\footnote{https://github.com/huggingface/transformers} and fine-tuned it. Adam optimizer was used and the learning rate was 2e-5, except when adversarial training on SIGHAN 13 dataset, which was 1e-5. We followed  \citet{zhang-etal-2020-spelling} to set our hyper-parameters. The size of the hidden state in Bi-GRU in Soft-Masked BERT was 256.

Similarly, we followed the hyper-parameters settings of  SpellGCN  \cite{cheng-etal-2020-spellgcn} except the batch size. Batch size was reduced to eight due to GPU memory. The BERT model used in SpellGCN was provided by the repository of BERT\footnote{https://github.com/google-research/bert}.

We conducted adversarial training on base models gained through pre-training and fine-tuning. The threshold $\lambda$ was tuned on the validation set for each dataset. The number of sentence pairs directly used for training was twice that that used to generate adversarial examples.

\subsection{Results and Analysis}

As shown in Table \ref{tab:performance}, through pre-training particularly designed for CSC, the models achieve better results on three benchmark datasets. 
The average improvement of correction F1 score was $4.3\%$ over base CSC models, which proves that our pre-training method has significant contribution to improving the model. Notably, BERT achieves state-of-the-art results on three datasets through our method.
\begin{figure}[H]
\centering 
\setlength{\textfloatsep}{2pt}
\setlength{\intextsep}{2pt}
\setlength{\abovecaptionskip}{2pt}
\includegraphics[width=3in]{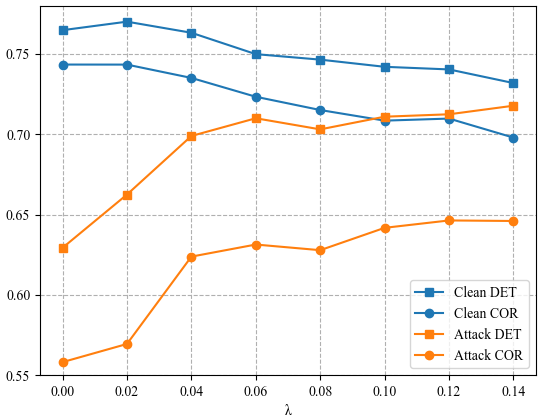}
\caption{Trade-off between generalization and robustness. The blue and orange lines respectively denote the average F1 scores of BERT on the SIGHAN-2015 data set and the adversarial examples generated ($\lambda = 0.05$).}\label{fig:1}
% \vspace{-3mm}
\end{figure}
Figure \ref{fig:1} shows the trade-off between generalization and robustness during adversarial training. As the threshold increases, the robustness of BERT also increases with a slight performance decrease on clean dataset (less than 0.7\%).

The experiments of the models under adversarial attacks were conducted with the base, pre-trained and adversarially trained models ($\lambda = 0.02$). 
We found that CSC models are vulnerable to adversarial examples as expected. 
The average drop in F1 score of three base models was $51.6\%$.
Under the attacks, the F1 scores of adversarially trained model decreased less ($44.1\%$), which indicates the adversarial training can substantially improve the robustness of CSC models. 
Compared with other models, BERT is more robust against adversarial attack (-41.2\%). 
The reason for the more serious robustness issues of other models may be related to the modules added to BERT, which increases the number of parameters, therefore it is more likely to overfit on the CSC data set.

\section{Conclusion}
% In this paper, we proposed the pre-training and adversarial training methods particularly for CSC models. Adversarial examples are generated by identifying vulnerable positions of a target CSC model and replacing these tokens with the spelling mistakes that most likely to occur in the contexts. Experimental results demonstrate that the performance and robustness of CSC models can be substantially improved through pre-training and adversarial training.

In this paper, we have described a character substitution-based method to create large pseudo data to pre-train the models by encouraging them to explore unseen misspellings. 
We also proposed a data augmentation method for training the CSC models by continually adding the adversarial examples, particularly generated to alleviate the weak spot of the current model, to the training set.
By the proposed pre-training strategy and adversarial training method, we can pursue both the exploration and exploitation when training the CSC models.
Experimental results demonstrate that the CSC models trained with the data augmented by these pseudo data and adversarial examples can substantially be improved in both generalization and robustness.

% \section{Appendix}

\section*{Acknowledgements}
The authors would like to thank the anonymous reviewers for their valuable comments. This work was supported by National Key R\&D Program of China (No. 2018YFC0830900), Shanghai Municipal Science and Technology Major Project (No. 2021SHZDZX0103), National Science Foundation of China (No. 62076068) and Zhangjiang Lab.

\bibliographystyle{acl_natbib}
\bibliography{acl2021}

%\appendix

\end{CJK}
\end{document}